\documentclass[10pt]{article} 
\usepackage[preprint]{tmlr}

\usepackage{amsmath,amsfonts,bm}

\def\eqref#1{equation~\ref{#1}}

\def\1{\bm{1}}

\DeclareMathAlphabet{\mathsfit}{\encodingdefault}{\sfdefault}{m}{sl}
\SetMathAlphabet{\mathsfit}{bold}{\encodingdefault}{\sfdefault}{bx}{n}

\usepackage{amsmath}
\usepackage{comment}
\usepackage{hyperref}
\usepackage{url}
\usepackage{booktabs}

\title{On the Design Space Between Transformers and Recursive Neural Nets}

\author{\name Jishnu Ray Chowdhury \email jishnu.ray.c@gmail.com \\
             \addr Computer Science Department\\
      University of Illinois at Chicago
      \AND
      \name Cornelia Caragea \email cornelia@uic.edu \\
      \addr Computer Science Department\\
      University of Illinois at Chicago}

\def\month{MM}  
\def\year{YYYY} 
\def\openreview{\url{https://openreview.net/forum?id=XXXX}} 

\begin{document}

\maketitle

\begin{abstract}
In this paper, we study two classes of models, Recursive Neural Networks (RvNNs) and Transformers,
 and show that a tight connection between them emerges from the recent development of two recent models - Continuous Recursive Neural Networks (CRvNN) and Neural Data Routers (NDR). On one hand, CRvNN pushes the boundaries of traditional RvNN, relaxing its discrete structure-wise composition and ends up with a Transformer-like structure. On the other hand, NDR constrains the original Transformer to induce better structural inductive bias, ending up with a model that is close to CRvNN. Both
 models, CRvNN and NDR, show strong performance in algorithmic tasks and generalization in which simpler forms of RvNNs and Transformers fail. We explore these “bridge” models in the design space between RvNNs and Transformers, formalize their tight connections, discuss their limitations, and propose ideas for future research. 
\end{abstract}

\section{Introduction}
\label{sec:missing-intro}
Consider a task like ListOps \citep{nangia2018listops} where we have to solve a nested list of mathematical operations such as: 
{\small{\verb+MAX(1,3,SUM(4,5,MIN(9,7)),4))+}}. Note that in such a task, unconstrained all-to-all (all tokens to all tokens) interactions cannot accommodate for fixed layer depth in Transformers \citep{vaswani2017attention} compared to RNNs where layer depth can extend adaptively based on sequence length. To solve this task, a model has to process the sequence sequentially because the operations in the outer lists can only be executed once the inner list operations are executed. Keeping that in mind, we can list a few potentially necessary {\em conditions} or requirements to fulfill for a task like this (and other tasks like program synthesis, program execution, mathematical reasoning, algorithmic reasoning, etc.): 

\begin{itemize}
    \item \textbf{C1: }Ability to operate in arbitrary orders (for example, the innermost operations may occur in any position in the input). 
    \item \textbf{C2: } Being equipped with a gating mechanism to keep some hidden states unchanged if needed (for example, outermost values may need to remain unchanged for a while to wait for inner list operations to be completed). 
    \item \textbf{C3: } Having an adaptive number of layers (according to the input) such that the number of layers can potentially increase indefinitely with higher demand (for example, when a higher depth of reasoning is required according to sequence length as longer sequences are more complex).
\end{itemize}

A similar list of requirements was also motivated in \citet{csordas2022the}. Transformers \citep{vaswani2017attention} with their fixed layers (failing C3) and lacking a gating mechanism to control information flow (failing C2) tend to struggle in tasks like ListOps \citep{tran2018importance,shen2019ordered,tay2021long,csordas2022the} especially in out-of-distribution contexts - for example, when evaluated on data of higher sequence length compared to the length of training samples. \citet{zhou2023algorithms} show that standard Transformers are most suited for length generalization when the problem can be solved by a short RASP-L program; failing otherwise. Even traditional RNNs without a sophisticated memory mechanism \citep{shen2019ordered} can struggle because their inductive bias of processing the sequence in a left-to-right manner can practically interfere with the C1 requirement above. Although ListOps is a simple task, paying attention to such `toy tasks' may hold the key to developing more robust architectures that can generalize productively and systematically \citep{csordas2022the,csordas2022ctl}---areas where even pre-trained language models can struggle \citep{kim2022uncontrolled}. Recently, a new variant of Transformer, Neural Data Router (NDR) \citep{csordas2022the} came close to solving ListOps by addressing some of the above requirements. On the other hand, our proposed Continuous Recursive Neural Networks (CRvNN) \citep{chowdhury2021modeling}, an extension of traditional Tree-RvNN, also shows exceptional length-generalization performance on ListOps without any user-specified structural information (learning everything from data). In this paper, we show that even though CRvNNs and NDRs are created from two different directions, a strong connection emerges between the two. In fact, they can be understood as `bridge models' (or `the missing links') between Tree-RvNNs and Transformers.

\section{General Schema}
\label{general_schema}
At the broadest level, both NDR (a variant of Transformer) and CRvNN (a variant of RvNN) can be formalized in terms of a recursive application of some function $Rec: \mathrm{I\!R}^{s \times d} \rightarrow \mathrm{I\!R}^{s \times d}$. The application of such a function at the recursive step $t$ (or equivalently, sharing weights at every layer)  can be written as:
\begin{equation}
    H^t, E^t = Rec(H^{t-1},E^{t-1})
    \label{general_recursion}
\end{equation}
In this chapter, we focus only on the encoder function (not the decoder). Here, $H^t, H^{t-1} \in \mathrm{I\!R}^{s \times d}$ represent a sequence of hidden states (e.g., $H^t = (H^t_1,H^t_2,\dots,H^t_s)$) where $s$ is the sequence length and $d$ is the dimension of hidden states. $E^t \in \mathrm{I\!R}^{s \times 1}$ represents a sequence mask. Normally, the sequence mask can be thought of as a mask for sequence padding\footnote{Typically used for efficient batched training.} with $0$ for pads and $1$ otherwise. However, in the case of CRvNNs, we can treat $E^t$ more generally as a ``soft sequence mask'' with continuous values in $[0,1]$. Going a little more granular, the general internal structure of $Rec$ can be formalized as:
\begin{equation}
    Rec(H^t, E^t) = Compose(Retrieve(H^t, E^t), H^t)
    \label{general_recursion2}
\end{equation}
The $Retrieve$ function can be thought to retrieve information for any state $H^t_i$ from the surrounding context $H^t$. The $Compose$ function can be thought to integrate the retrieved information with the original $H^t_i$. Roughly, we will show some analogous functions that fulfill these roles in both NDRs and CRvNNs. In $\S$\ref{vanilla_connection}, we also discuss how this recursive formalism can also represent vanilla Transformers \citep{vaswani2017attention}. 

Next, we show how NDR and CRvNN can be taken as specific instances of this general recursive schema as defined in Eq. \ref{general_recursion2} and how a strong similarity appears between NDR and CRvNN despite being created from two different perspectives.

\section{Neural Data Router (NDR)}
\label{ndr}
Neural Data Routers \citep{csordas2022the} can be construed as a modified version of Universal Transformers (UTs) \citep{dehghani2018universal} that similarly employs layer-wise parameter sharing or recursion (repeats the same layer function $Rec$). 

\noindent \textbf{Formalism of Retrieve Function}: The $Retrieve$ function of NDR can be said to constitute the Multi-headed Attention Function (MHA) \citep{vaswani2017attention}: 
\begin{equation}
    X^t, E^t = MHA(H^t, E^t) 
\end{equation}
For ease of explanation, we concentrate on the single-headed version (although the multi-headed version is used in the experiments). Inside the single-headed attention of NDR, there is a non-traditional form of attention called geometric attention:
\begin{equation}
Q = H^t W_Q, \;K = H^t W_K, \;V = H^t W_V
\end{equation}
\begin{equation}
    C = sigmoid(Q K^T) \odot E^t
    \label{match}
\end{equation}
\begin{equation}
    A_{ij} = C_{ij} \Pi_{k \in \mathbb{S}_{ij}} (1-C_{ik})
    \label{geometric}
\end{equation}
Here, $W_Q, W_K, W_V \in \mathrm{I\!R}^{d \times d_h}$, $H^t \in \mathrm{I\!R}^{s \times d}$, and $Q,K,V \in \mathrm{I\!R}^{s \times d_h}$. $A \in {\rm I\!R}^{s \times s}$ is the attention matrix where $A_{ij}$ represents how much $H^t_i$ attends to $H^t_j$. $\mathbb{S}_{ij}$ contains all positions $k$ except $i$ and $j$ such that $|i-k| < |j-i|$ if $i < j$ else  $|i-k| \leq |j-i|$ if $i \geq j$. Additionally, the diagonals of $A_{ii}$ are set as $0$. Plainly said, geometric attention makes each query prefer to attend to the \textit{closest matching} values (closest in terms of relative distances), suppressing more distant attention proportionately. This is motivated by the fact that algorithmic tasks often benefit from a preference for local operations. Finally, we have:
\begin{equation}
    Z = A_{ij} V,\; Y = Z W_o
\end{equation}
\begin{equation}
    X^t = LN_1(Y + H^t)
\end{equation}
Here, $W_o \in {\rm I\!R}^{d_h \times d}$ and $LN_1$ is layer norm. NDR \citep{csordas2022the} also introduces a mechanism to learn to reduce attention to positions in a specific direction, but currently, we ignore these implementational details for focus. 
 
\noindent \textbf{Formalism of Compose Function}:
In the $Compose$ function, NDR introduces a gating mechanism: 
\begin{equation}
    G = sigmoid(FFN_{gate}(X^t)) 
\end{equation}
 $FFN_{gate}: {\rm I\!R}^{s \times d} \rightarrow {\rm I\!R}^{s \times d}$ can be any multi-layered position-wise feedforward network, and $G \in {\rm I\!R}^{s \times d}$ are the gates (multidimensional). The output of the $Compose$ function is then:
\begin{equation}
    H^{t+1} = G \odot LN_2(FFN_{data}(X^t)) + (1-G) \odot H^t
    \label{gate_1}
\end{equation}
\begin{equation}
    E^{t+1} = E^t
\end{equation}
Here $LN_2$ is layer norm and $FFN_{data}: {\rm I\!R}^{s \times d} \rightarrow {\rm I\!R}^{s \times d}$ is another multi-layered position-wise feedforward network. 

\noindent \textbf{Requirements Satisfaction in NDR:} 
NDRs \citep{csordas2022the} address many of the desiderata described in the introduction. NDR addresses C1 by using a form of self-attention to process data in any arbitrary order, but unlike Transformers, it uses geometric attention to process data in a more controlled fashion. NDR explicitly addresses C2 by incorporating a gating mechanism in Eq. \ref{gate_1} (although, note that certain other Transformers also incorporate a gating mechanism \cite{chai2020highway}). NDR partially addresses C3 by sticking to sharing parameters across layers, similar to Universal Transformers. Thus, it allows us to increase the number of layers during inference indefinitely. However, NDR remains inflexible because it cannot completely adapt according to the input. In the original work, the expected maximum depth has to be considered a priori to set the hyperparameters for the maximum depth of layers during training \cite{csordas2022the}. On the other hand, although we can use the sequence size as a heuristic for maximum layer depth without dynamic halting, the model can become exorbitantly expensive if done so because it lacks any early halting mechanism.

\section{Continuous Recursive Neural Network (CRvNN)}
We already showed CRvNNs to be a continuous relaxation of traditional Tree-RvNNs. Here, we show that CRvNNs can also, at the same time, be construed as a form of a constrained Transformer (particularly, a constrained NDR). 

\noindent \textbf{Formalism of Retrieve Function}: The $Retrieve$ function of CRvNN can be said to constitute an attention-like function (say, $NR$ - neighbor retriever) to retrieve an immediate left neighbor: 
\begin{equation}
    X^t, E^t = NR(H^t, E^t) 
\end{equation}
Inside $NR$, we have:
\begin{equation}
    A_{ij} = E_j \Pi_{k \in \mathbb{S}_{ij}} (1-E_{k})
    \label{geometric2}
\end{equation}
\begin{equation}
     X^t = A H^t
\end{equation}
Here, the set $\mathbb{S}_{ij}$ contains all positions $k$ except $i$ and $j$ such that $i-k < i-j$ and $i-k > 0$. $A \in {\rm I\!R}^{s \times s}$ acts like an attention matrix and $A_{ij}$ is $0$ if $\mathbb{S}_{ij}$ is empty. Eqn. \ref{geometric2} is a reformulation of Eqn. $6$ in $\S 3.2.1$ from \citet{chowdhury2021modeling}. It is one of the two variants proposed for modeling $A_{ij}$ in $\S 3.2.1$ (denoted as $P_{ij}$ in $\S$$\S 3.2.1$) in \citet{chowdhury2021modeling}.

A unique aspect of CRvNN is that it allows for the soft deletion of processed information so that it does not confuse future recursive steps. For example once \verb+MAX(1,3,SUM(4,5,MIN(9,7)),4))+ is simplified to \verb+MAX(1,3,SUM(4,5,7),4))+, we do not need to keep around information related to $MIN(9,7)$ explicitly. The information in the corresponding positions can be `deleted.' This is modeled by $E^t$ (initialized as the pad mask) where $E^t_i$ represents the probability of position $i$ having not been deleted so far (alternatively, we may say $E^t_i$ is the existential probability of position $i$ to use the vocabulary we used in \citet{chowdhury2021modeling}). Now, we can see that $A_{ij}$ represents the probability that the item in $j$ is the first existing item left to $i$. Thus, $X^t$ represents the softly retrieved left neighbors of $H^t$. 

\noindent \textbf{Formalism of Compose Function}: 
The $Compose$ function can be described as:
\begin{equation}
    G = DF(H^t, E^t), \; L = A G
\end{equation}
\begin{equation}
    H^{t+1} = L \odot \text{Cell}(X^t,H^t) + (1-L) \odot H^t
    \label{gate2}
\end{equation}
\begin{equation}
    E^{t+1} = E^t \odot (1-G)
    \label{deletion}
\end{equation}
Here $DF: {\rm I\!R}^{s \times d} \times {\rm I\!R}^{s \times 1} \rightarrow {\rm I\!R}^{s \times 1}$ is some arbitrary function which we can refer to as the decision function serving a similar role as $FFN_{gate}$ in NDR. $G,L \in {\rm I\!R}^{s \times 1}$  and $Cell: {\rm I\!R}^{s \times d} \times {\rm I\!R}^{s \times d} \rightarrow {\rm I\!R}^{s \times d}$ can be any arbitrary recursive cell function like an LSTM \citep{hochreiter1997long} or the Transformer-inspired Gated Recursive Cell (GRC) \citep{shen2019ordered}.  

We can interpret $G_i$ as the probability for the information in $i$ to be pushed leftwards (deleting it from the current position)\footnote{We called it as the composition probability in \cite{chowdhury2021modeling}}. This interpretation provides intuition for Eq. \ref{gate2} and Eq. \ref{deletion}. Given this interpretation, $L_i$ (where $L = A G$) can represent the probability of pulling some information from rightwards. This explains Eq. \ref{gate2}, where we update a position $i$ based on its $L_i$ score, which indicates if new information is being pulled. Moreover, $G_i$ can also be simultaneously treated as a deletion probability (if the information in $i$ is pushed leftwards, position $i$ itself can be deleted). Thus, $E^{t+1}$ is updated based on $G$ as in Eq. \ref{deletion} to reduce existential probability according to the current deletion probability. 

\noindent \textbf{Requirements Satisfaction in CRvNN:} Like a Transformer, CRvNN can also process sequences in arbitrary orders satisfying C1. It has a gating mechanism satisfying C2 and a dynamic halting mechanism satisfying C3. However, although CRvNN satisfies these requirements, it does so at the cost of stronger inductive bias and less flexibility. Unlike CRvNN, in principle, NDRs can model more general non-projective structures for information flow.

\section{Connections between CRvNN and NDR}
With our new formalism, the connection between NDR and CRvNN (and by extension the connection between Transformers and Tree-RvNNs) becomes much more explicit (see Table \ref{table:similarities}). In the $Retrieve$ component, note how Eq. \ref{geometric2} to create the attention-like matrix in CRvNN turns out to be of the same form as Eq. \ref{geometric} for geometric attention in NDR. The main difference is that CRvNN does not use query-key interaction\footnote{However, the attention in CRvNN still indirectly depends on gates predicted by $DF$ that can allow any arbitrary interaction depending on the particular implementation.} and the set $\mathbb{S}_{ij}$ is defined differently in both cases. Particularly given the restrictions in $\mathbb{S}_{ij}$, the attention in CRvNN can be said to be a strictly \textit{directionally masked} variation of geometric attention from NDR. Beyond that, in the $Compose$ component, we again find a striking similarity between \ref{gate2} for CRvNN and \ref{gate_1} from NDR. Both represent a gating mechanism to control information updates. Besides the similarities, a few differences worth emphasizing include that CRvNN lacks any multi-head setup, and its gates are scalar.

\begin{table*}[t]
\small
\centering
\def\arraystretch{1.2}
\begin{tabular}{ l | l | l } 
\hline
\textbf{Component} & \textbf{NDR} & \textbf{CRvNN}\\
\hline
Attention & $A_{ij} = C_{ij} \Pi_{k \in \mathbb{S}_{ij}} (1-C_{ik})$ & $A_{ij} = E_j \Pi_{k \in \mathbb{S}_{ij}} (1-E_{k})$\\
Gating & $H^{t+1} = G \odot LN_2(FFN_{data}(X^t)) + (1-G) \odot H^t$ &  $H^{t+1} = L \odot \text{Cell}(X^t,H^t) + (1-L) \odot H^t$\\
\hline
\end{tabular}
\caption{Connection between NDR and CRvNN.}.
\label{table:similarities}
\end{table*}

\vspace{1mm}
\noindent \textbf{Dynamic Halt}: One of the most critical differences between NDRs and CRvNNs is dynamic halt. CRvNN's maintenance of existential probabilities makes up for a convenient mechanism for a dynamic halt. First, we can use a threshold to convert the probabilities into binary values, and then we can choose to halt whenever all but one position exists with a binarized value of $1$ according to the existential probability. In contrast, NDR lacks any form of dynamic halt (see the end of \S\ref{ndr}). 

\vspace{1mm}

\section{Transformers under the Recursive Formalism}
\label{vanilla_connection}
Here, we explain how the vanilla Transformer can fit under the recursive formulation introduced in $\S$\ref{general_schema}. First, note that the recursive formalism in $\S$\ref{general_schema}  does at least represent Universal Transformer (UT) \citep{dehghani2018universal}. UT repeatedly applies the same function with the same transformer weights in every layer during encoding (equivalent to sharing parameters in every layer). Second, it can also be shown that any vanilla Transformers with `unshared parameters' can be re-expressed in the form of a UT with recursive layers and repeated parameters by increasing the hidden state size \citep{bai2019deep} (See the supplementary C in \citet{bai2019deep}). Thus, by extension, the general schema in $\S$\ref{general_schema} also accommodates vanilla Transformers.

\section{NDR Empirical Analysis}
We performed some empirical experiments with NDR on two datasets, ListOps and Logical Inference, to see where it stands compared to CRvNN.

\label{sec:ndr_analysis}
\subsection{ListOps-DG2}
\noindent \textbf{Dataset Settings: } ListOps or List Operations is an arithmetic task for solving a nested list of mathematical operations as exemplified in $\S$\ref{sec:missing-intro}. ListOps-DG is a special split originally set up in \citet{Chowdhury2023beam} to test depth-generalization capacities of models. Depth of a ListOps sequence represents the maximum number of nested operators present in it. The training set of ListOps-DG has $\leq$ $6$ depths, but the test split (DG split) has higher depths ($8$-$10$). There are also other splits for length generalization, higher depth generalization, and argument generalization. The test split from Long Range Arena (LRA) \cite{tay2021long} version of ListOps is also used. Details are present in \citet{Chowdhury2023beam}, and the main data parameters per split are presented in Table \ref{table:ndr_listopsdg}. ListOps-DG2 - the exact data used here - is designed to be similar to ListOps-DG. ListOps-DG2 was also originally presented in \citet{Chowdhury2023beam}. The main difference between ListOps-DG and ListOps-DG2 is that ListOps-DG2 has 1 million training samples, which is $\sim 10$ times more than that in ListOps-DG. We did this because originally, NDR was also trained in a similar sized ListOps dataset \citep{csordas2022the}. The training samples have $\leq 100$ sequence length, $\leq 5$ arguments, and $\leq 6$ depth (depth is the maximum number of nested operators). 

\begin{table*}[t]
\small
\centering
\def\arraystretch{1.2}
\begin{tabular}{  l | l | l l l | l l |l} 
\hline
\textbf{Model} & \textbf{DG} & \multicolumn{3}{c}{\textbf{Length Gen.}} & \multicolumn{2}{|c|}{\textbf{Argument Gen.}} & \textbf{LRA}\\
(Lengths) & $\leq$ 100 & 200-300 & 500-600 & 900-1k & 100-1k & 100-1k & 2K\\
(Arguments) & $\leq$ 5 & $\leq$ 5 & $\leq$ 5 & $\leq$ 5 & 10 & 15 & $\leq$ 10\\
(Depths) & 8-10 & $\leq$ 20 & $\leq$ 20 & $\leq$ 20 & $\leq$ 10 & $\leq$ 10 & $\leq$ 10\\
\hline
Transformer & $63.24$ & $19$ & $11.15$ & $10.1$ & $19.2$ & $15.75$ & $10.35$\\
NDR (24 layers) & $95.1$ & $33.25$ & $25.6$ & $18.2$ & $67.75$ & $55.55$ & $44.3$\\
NDR (48 layers) & $91.25$ & $28.60$ & $15.95$ & $13.8$ & $\mathbf{73.1}$ & $\mathbf{61.19}$ & $46.1$\\
CRvNN & $\mathbf{100}$ & $\mathbf{99.9}$ & $\mathbf{99.8}$ & $\mathbf{99.4}$ & $67.10$ & $46.05$ & $\mathbf{66.35}$\\
\bottomrule
\end{tabular}
\caption{We report the accuracy (median of 3 runs) on a ListOps-DG2. The models were trained on samples with lengths $\leq$ 100, depth $\leq$ 6, and arguments $\leq$ 5 and then tested on OOD splits with higher depth (DG split), higher lengths, splits with higher arguments (argument gen. splits), and the test set of LRA.}
\label{table:ndr_listopsdg}
\end{table*}

\begin{table*}[t]
\small
    \centering
    \small
    \begin{tabular}{l cccccc}
    \toprule
    \textbf{Model} & \multicolumn{6}{c}{\textbf{Number of Operations}}\\
          & 7 & 8 & 9 & 10 & 11 & 12 \\
    \midrule
    Transformer & $90.65$ & $82.34$ & $73.27$ & $67.45$ & $57.41$ & $52.29$\\
    NDR (15 layers) & $93.63$ & $89.33$ & $87.27$ & $81.16$ & $78.47$ & $73.85$\\
    NDR (30 layers) & $90.8$ & $86.17$ & $84.75$ & $78.12$ & $75.46$ & $72.57$\\
    CRvNN & $\mathbf{97.79}$ & $\mathbf{97.13}$ & $\mathbf{96.1}$ & $\mathbf{94.59}$ & $\mathbf{94.44}$ & $\mathbf{92.15}$\\
    \bottomrule
    \end{tabular}
        \caption{
    Test accuracy of models trained on samples with $\leq 6$ logical operators and tested on out-of-distribution samples (number of logical operators between 7-12). We bold the best results per block separately. We show the median of 3 different runs.} 
    \label{table:ndr_proplogic}
\end{table*}

\noindent \textbf{Model Settings:} We compare three models - a relatively vanilla Transformer, NDR, and CRvNN. For the NDR, we use the same hyperparameters as used for ListOps by Csordas et al. \citep{csordas2022the}. For CRvNN, we use the same hyperparameters as we used before for ListOps-DG in \citet{Chowdhury2023beam}. For the baseline Transformer, we use FlashAttention2 \citep{dao2024flashattention} and xPos as positional encoding \citep{sun2023length}. We train NDR with $20$ layers. We used the same hyperparameters for Transformer as used for ListOps in the original work presenting Long Range Arena \citep{tay2021long}.  

\noindent \textbf{Results: } We present the results in Table \ref{table:ndr_listopsdg}. As we can see from the results, the Transformer baseline does not perform as well. NDR, with better inductive biases, performs quite well and generalizes to slightly higher depths (the DG split) consistent with prior results \citep{csordas2022the}. However, it fails to generalize to much higher lengths and depths. Here, in contrast, CRvNN performs much better and achieves $\geq 90\%$ accuracy in all length and depth generalization splits. Interestingly, NDR still outperforms CRvNN on argument generalization. 

Following the suggestions of \citet{csordas2022the}, we also increased the number of layers during inference (e.g., up to $24$ and $48$ layers) to handle higher-depth sequences. Increasing the layer number during inference is possible because NDR shares its parameters layer-wise. Nevertheless, layer increases did not help - rather, they tended to harm the performance in our experiments. 

In summary, even after experiencing more data (from ListOps-DG2), NDR generalizes worse than how OM, CRvNN, or BT-GRC generalizes on ListOps-DG (as shown in \citet{Chowdhury2023beam}). Moreover, NDR requires some prior estimation of the true computation depth of the task for its hyperparameter setup for efficient training, unlike the other latent tree models.

\subsection{Logical Inference}
\noindent \textbf{Dataset Settings: } We also perform a few experiments of the logical inference dataset \citep{bowman2015tree}. Logical Inference \citep{bowman2015tree} is another structure-sensitive task where Transformers struggle to generalize \citep{tran2018importance}. In this task, the model has to classify the logical relationship between two given sequences in propositional logic formalism. For this task, the models are trained on samples with $\leq 6$ operators and then tested on samples with a higher number of logical operators ($7-12$). 

\noindent \textbf{Model Settings:} Again, we compare three models - a relatively vanilla Transformer, NDR, and CRvNN. We share the hyperparameters between ListOps-DG2 and Logical Inference for both CRvNN and NDR. One difference is that NDR was trained with $15$ layers since the training samples have lower tree depth. For the baseline Transformer, we use FlashAttention2 \citep{dao2024flashattention} and xPos as positional encoding \citep{sun2023length} as before. We used the same hyperparameters for Transformer as we used for logical inference in \citet{chowdhury2024recurrent}.

\noindent \textbf{Results: }  We present the results in Table \ref{table:ndr_proplogic}. We observe the same patterns here as we did in ListOps, except that NDR performs relatively better in generalization to a higher number of operators - perhaps because the sequence lengths of logical inference in test sets are still on the lower side (whereas in the case of ListOps, it can go in the range of 500-1000). Nevertheless, like before, NDR outperformed Transformer, but CRvNN still performed much better than NDR.

\section{Discussion and Future Directions}

\noindent \textbf{Structural Flexibility:} NDRs have a more flexible inductive bias than CRvNN in the retriever function because NDRs use key-value attention to attend to any of the closest matching elements. In principle, NDR could model non-projective or more general structures for information flow. However, in CRvNN, the retriever only allows attention to the closest existing element, which limits it to only approximate information flow through a projective constituency structure. While the stronger inductive bias in CRvNN can be helpful, it can also limit its scalability to more complex tasks when more data is available \citep{sutton2019bitter}. At the same time, some degree of inductive bias is, arguably, still necessary for OOD generalization \citep{mitchell1980need,goyal2022inductive}. The ideal is to find a sweet spot where the vicinity of the design space explored here may be a promising region to investigate.

\noindent \textbf{Permanent Deletion:} CRvNN enforces a more controlled information flow compared to NDRs. One interesting distinction that allows this is that CRvNN uses a form of permanent (soft) deletion of token positions through the existential probabilities (which are monotonically decreasing per recursive function). This can allow for a better inductive bias for ignoring processed information in a consistent manner. 

\noindent \textbf{Dynamic Halting:} One immediate future direction can be to investigate different ways to incorporate dynamic halt mechanisms \citep{schmidhuber2012delimiting,graves2016adaptive,banino2021ponder} in NDR. Another alternative in a similar vein can be to turn NDR into a Deep Equilibrium Network \citep{bai2019deep}, which can implicitly and adaptively increase recursive iterations based on sample difficulty. Promisingly, a recent work even shows that Deep Equilibrium models can also help in OOD length generalization in certain synthetic tasks \citep{liang2021out}. As a plus, Deep Equilibrium models can also significantly reduce the memory consumption of these models \citep{bai2019deep}. We also explore some novel halting mechanisms in \cite{chowdhury2024recurrent}. This could be further investigated with NDR or adjacent Transformer models.

\bibliography{main}
\bibliographystyle{tmlr}

\end{document}